\documentclass{article} 
\usepackage[preprint]{colm2026_conference}
\usepackage{algorithm}
\usepackage[T1]{fontenc}
\usepackage{upquote}
\usepackage{algorithmic}
\usepackage{microtype}
\usepackage{hyperref}
\usepackage{url}
\usepackage{booktabs}
\usepackage{amssymb}
\usepackage{afterpage}
\usepackage{makecell}
\usepackage{graphicx}
\usepackage{wrapfig}
\usepackage[most]{tcolorbox}
\usepackage{listings}
\makeatletter
\newenvironment{breakablealgorithm}
  {
   \begin{center}
     \refstepcounter{algorithm}
     \hrule height.8pt depth0pt \kern2pt
     \renewcommand{\caption}[2][\relax]{%
       {\raggedright\textbf{Algorithm~\thealgorithm} ##2\par}
       \ifx\relax##1\relax\else
         \addcontentsline{loa}{algorithm}{\protect\numberline{\thealgorithm}##1}
       \fi
       \kern2pt\hrule\kern2pt
     }
  }{
     \kern2pt\hrule\relax
   \end{center}
  }
\makeatother

\usepackage{lineno}
\usepackage{amsmath}
\definecolor{darkblue}{rgb}{0, 0, 0.5}
\hypersetup{colorlinks=true, citecolor=darkblue, linkcolor=darkblue, urlcolor=darkblue}

\title{Aligning Progress and Feasibility: A Neuro-Symbolic Dual Memory Framework for Long-Horizon LLM Agents}

\author{Bin Wen$^{1}$, Ruoxuan Zhang$^{2}$, Yang Chen$^{1}$, Hongxia Xie$^{2}$, Lan-Zhe Guo$^{1}$\thanks{Corresponding author}
\\
$^1$Nanjing University \quad $^2$Jilin University
}

\begin{document}

\ifcolmsubmission
\linenumbers
\fi

\maketitle

\vspace{-3mm}
\begin{abstract}

   \vspace{-2mm}
   \noindent %
Large language models (LLMs) have demonstrated strong potential in long-horizon decision-making tasks, such as embodied manipulation and web interaction. However, agents frequently struggle with endless trial-and-error loops or deviate from the main objective in complex environments.
We attribute these failures to two fundamental errors: \emph{global Progress Drift} and \emph{local Feasibility Violation}. Existing methods typically attempt to address both issues simultaneously using a single paradigm. However, these two challenges are fundamentally distinct: the former relies on fuzzy semantic planning, while the latter demands strict logical constraints and state validation. The inherent limitations of such a single-paradigm approach pose a fundamental challenge for existing models in handling long-horizon tasks.
Motivated by this insight, we propose a Neuro-Symbolic Dual Memory Framework that explicitly decouples semantic progress guidance from logical feasibility verification. Specifically, during the inference phase, the framework invokes both memory mechanisms synchronously: on one hand, a neural-network-based Progress Memory extracts semantic blueprints from successful trajectories to guide global task advancement; on the other hand, a symbolic-logic-based Feasibility Memory utilizes executable Python verification functions synthesized from failed transitions to perform strict logical validation.
Experiments demonstrate that this method significantly outperforms existing competitive baselines on ALFWorld, WebShop, and TextCraft, while drastically reducing the invalid action rate and average trajectory length.

\end{abstract}

\vspace{-4mm}
\section{Introduction}
\label{sec:intro}

\afterpage{
\begin{figure}[!t]
    \centering
    \includegraphics[width=0.98\linewidth]{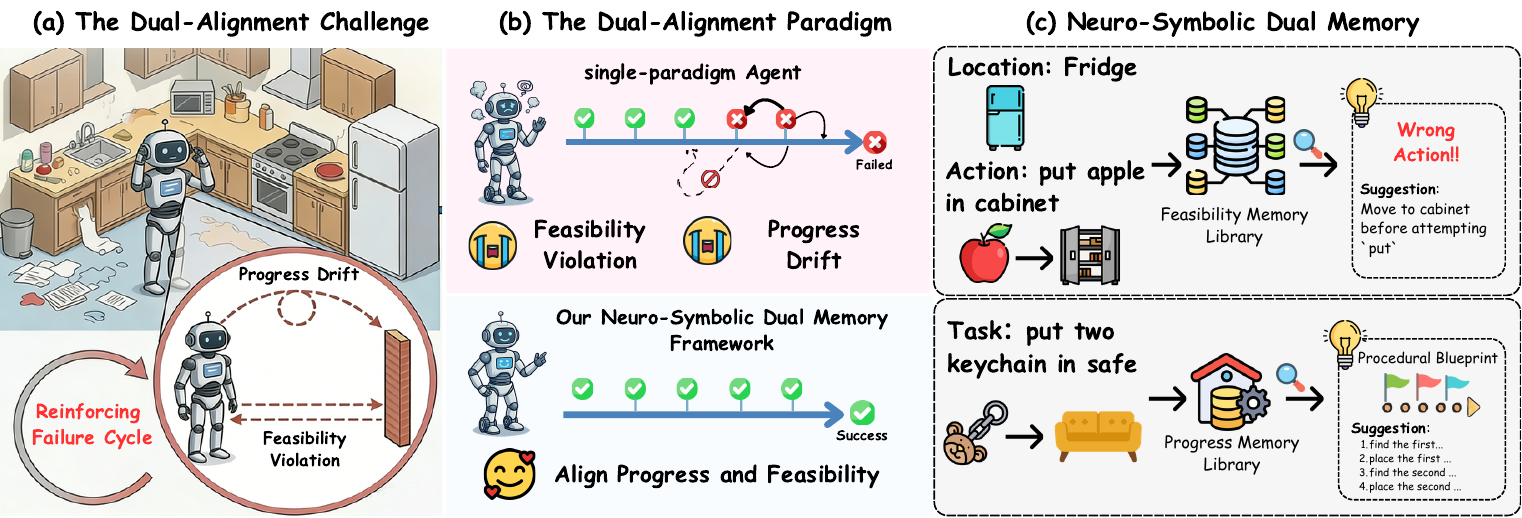}
    \caption{\textbf{Illustration of our neuro-symbolic dual-alignment framework.} \textbf{(a) The Dual-Alignment Challenge:} Long-horizon agents often trap themselves in a reinforcing failure cycle caused by coupled progress drift and feasibility failures. \textbf{(b) The Dual-Alignment Paradigm:} Our approach shifts from error-prone unaligned execution to a stable, dually aligned reasoning loop. \textbf{(c) Neuro-Symbolic Dual Memory:} The agent concurrently aligns local actions via executable symbolic rules in Feasibility Memory and anchors global steps via retrieved procedural blueprints in Progress Memory.}
    \label{fig:placeholder}
    \vspace{-2mm}
\end{figure}
}

In recent years, large language models (LLMs) have demonstrated immense potential as agents in tasks such as embodied manipulation and web interaction \citep{react,code,statler,webarena,mind2web,webvoyager,eureka,osworld,appagent}. However, when faced with complex environments characterized by strict action constraints and long-horizon dependencies, agents remain highly prone to inefficient trial-and-error or deviation from the main task objective \citep{travelplanner,planning}. For example, an embodied agent may repeatedly attempt to place an object into a receptacle without satisfying the required preconditions, while a web agent may drift across irrelevant products or filters without making progress toward the requested purchase. We attribute such failures in long-horizon control to two intertwined fundamental crises: \emph{Progress Drift} from a global perspective and \emph{Feasibility Violations} from a local perspective.

To address these challenges, most existing methods attempt to solve both problems simultaneously through a single framework or a unified experience representation \citep{reflexion,re2,expel,walle2,automanual,nesyc}. The core limitation of these methods is that they do not explicitly separate semantic progress guidance from logical feasibility verification, even though progress alignment and feasibility alignment impose fundamentally different requirements. Global progress is inherently fuzzy and context-dependent, requiring high-dimensional semantic matching and generalization from successful historical experiences \citep{memory}; whereas the feasibility of local actions is determined by the physical laws of the environment, requiring absolutely strict logical boundaries and conditional triggers \citep{walle2,llmp}. Forcing these two into a single paradigm often causes neural networks to hallucinate when confronted with hard constraints \citep{planbench,position}, or leaves symbolic rules lacking the flexibility needed to handle complex semantics \citep{neuro,novelty}.

Building on this insight, we argue that resolving the dual-alignment crisis requires matching each alignment objective with an architecture tailored to the type of reasoning it demands.
 Specifically, Progress Alignment relies on semantic generalization and is thus best modeled using neural mechanisms; conversely, Feasibility Alignment relies on rigorous logical validation and is therefore best constrained by symbolic mechanisms. The essence of long-horizon agent tasks is precisely the organic integration of semantic planning and logical constraints.

To this end, we propose the Neuro-Symbolic Dual Memory Framework, as illustrated in Figure \ref{fig:placeholder}. This framework explicitly decouples these two capabilities within a unified inference loop. On the one hand, we design a neural-based Progress Memory, which transforms successful historical trajectories into semantic blueprints with progress anchors to guide the agent's global advancement. On the other hand, we introduce a symbolic-based Feasibility Memory, which distills executable Python code validators from failure transitions to perform strict hard-logic interception and precondition checks before the agent submits an action. This design ensures that the agent can maintain a clear global vision while securing stable local execution.

Our contributions are as follows.
\vspace{-1mm}
\begin{itemize}
    \item \textbf{Dual-alignment view of long-horizon failure.} We identify long-horizon agent failure as arising from two coupled but distinct objectives: global progress alignment and local feasibility alignment. This view explains why one mechanism is often insufficient for both semantic progress and strict feasibility, motivating neural guidance for the former and symbolic verification for the latter.

    \item \textbf{Neuro-symbolic dual memory framework.} We propose a Neuro-Symbolic Dual Memory Framework that instantiates this view with a neural Progress Memory for stage-aware semantic guidance and a symbolic Feasibility Memory for executable action verification within a unified inference loop.

    \item \textbf{Extensive experiments.} We evaluate the framework on three representative long-horizon benchmarks, ALFWorld, WebShop, and TextCraft. Our method consistently outperforms strong baselines, and ablations further show that Progress Memory mainly improves stage-level advancement whereas Feasibility Memory mainly reduces invalid actions, supporting the complementarity of the two modules.
\end{itemize}

\section{Related Work}
\label{sec:related}

\paragraph{LLM Agents.} Large language models are now the standard backbone for long-horizon agents, where they need to sustain multi-step planning, adapt to feedback, and maintain consistency over extended interaction sequences \citep{react,language,llmp,code,progprompt}. To improve robustness, prior work augments LLM agents with hierarchical decomposition, retrieval, state tracking, workflow memory, and experience-driven self-improvement \citep{adapt,voyager,yoo2024exploratory,stateact,awm,reflexion,expel,msiagent,automanual}. Despite their differences, these methods largely reuse trajectories, reflections, stage cues, and corrective heuristics within a shared semantic memory space through prompting, retrieval, or textual reflection. This paradigm is effective for high-level progress guidance, but it remains fundamentally based on fuzzy neural generalization. As a result, when the same representation is asked to handle local feasibility violations that require strict logical boundaries, it often becomes unreliable \citep{planbench,position}. 

\vspace{-1mm}
\paragraph{Neuro-Symbolic Agents.} Another line of work stabilizes agent behavior by grounding decision-making in explicit constraints, structured world knowledge, or neuro-symbolic control mechanisms \citep{saycan,sayplan,inner_monologue,walle2,re2,nesyc}. SayCan \citep{saycan} and SayPlan \citep{sayplan} combine language planning with affordance-aware grounding. More recent methods such as $Re^2$ Agent \citep{re2} and WALL-E 2.0 \citep{walle2} further reduce invalid actions through failure abstractions, action rules, and structured scene representations. While such designs help ensure action feasibility and environmental grounding, they often lack the flexibility required for complex and highly variable long-horizon tasks because control remains constrained by relatively rigid rule frameworks \citep{neuro,novelty}. In contrast, our framework explicitly separates semantic progress guidance from symbolic feasibility verification, allowing neural memory to handle global task advancement while symbolic memory enforces strict local executability.

\vspace{-2mm}
\section{Method}
\label{sec:method}

\vspace{-1mm}
\subsection{Overview}
\begin{figure*}[t!]

\centering
    \includegraphics[width=0.9\textwidth]{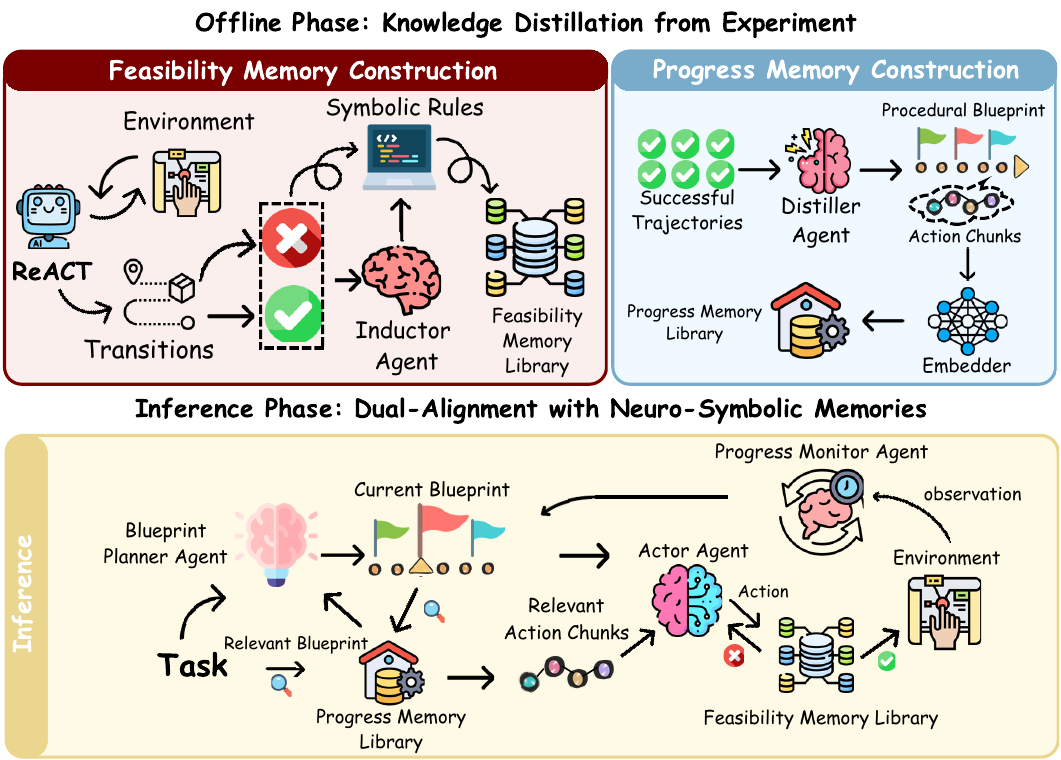}

\caption{\textbf{Overview of our neuro-symbolic dual memory framework.} The proposed system explicitly separates local feasibility alignment from global progress alignment according to the distinct reasoning demands of the two objectives. \textbf{Top (Offline Phase):} Failed interactions are compiled into executable symbolic verifier rules to construct the symbolic Feasibility Memory, while successful trajectories are distilled into stage-anchored procedural blueprints to form the neural Progress Memory. \textbf{Bottom (Inference Phase):} At inference time, Progress Memory provides stage-aware guidance for proposing progress-consistent actions, while Feasibility Memory performs symbolic feasibility verification and iterative refinement before execution.}

\label{fig:method}

\end{figure*}

In long-horizon tasks, the agent must simultaneously avoid locally infeasible actions and maintain global progress. We therefore explicitly decouple these two objectives with a neuro-symbolic dual memory design: a symbolic Feasibility Memory $\mathcal{M}_f$ induced from failed transitions for executable action verification, and a neural Progress Memory $\mathcal{M}_p$ built from successful trajectories for stage-aware semantic guidance. During inference, Progress Memory proposes progress-consistent actions, while Feasibility Memory verifies and refines them before execution, yielding a unified decision loop with decoupled knowledge representations. The overall pipeline is illustrated in Figure \ref{fig:method}.

Formally, we model the environment as a partially observable Markov decision process (POMDP), $\mathcal{E}=(\mathcal{S},\mathcal{A},\mathcal{O},\mathcal{T},\Omega)$, where $\mathcal{S}$, $\mathcal{A}$, and $\mathcal{O}$ denote the latent state, action, and observation spaces, $\mathcal{T}(s_{t+1}\mid s_t,a_t)$ is the transition function, and $\Omega(o_t \mid s_t)$ is the observation function. At step $t$, the agent selects an action according to $\pi(a_t \mid h_t,\mathcal{M}_p,\mathcal{M}_f)$, where $h_t$ denotes the interaction history, and $\mathcal{M}_p$ and $\mathcal{M}_f$ denote Progress Memory and Feasibility Memory. Finally, when the episode ends, the environment returns a binary reward $R(\tau)\in[0,1]$ indicating whether the task is completed. Our objective is:

\begin{equation}
\max_{\pi}\mathbb{E}_{\tau\sim\pi(\cdot\mid\mathcal{M}_p,\mathcal{M}_f)}\left[R(\tau)\right].
\end{equation}

To construct the dual memories, we first collect trajectories through the online interaction of a base agent. Specifically, we employ the ReAct \citep{react} strategy to explore 50 training tasks that are fully disjoint from the test set, yielding a trajectory dataset:
\begin{equation}
    \mathcal{D} = \left\{ \left(x^{(i)}, o_0^{(i)}, a_1^{(i)}, o_1^{(i)}, \dots, a_{T_i}^{(i)}, o_{T_i}^{(i)}, y^{(i)}\right) \right\}_{i=1}^{50}
\end{equation}

Next, Section \ref{feasibility} describes how symbolic Feasibility Memory is induced from failed transitions, Section \ref{progress} introduces how neural Progress Memory is built from successful trajectories, and Section \ref{inference} presents the unified dual-alignment inference loop powered by their synergy.

\subsection{Feasibility Alignment via Symbolic Memory}
\label{feasibility}

The objective of Feasibility Memory is to prevent local Feasibility Violations by enforcing action executability boundaries. In long-horizon tasks, many failures arise not from incorrect high-level planning, but from violating fine-grained preconditions imposed by the environment. Since feasibility alignment depends on strict, condition-triggered validation rather than fuzzy semantic generalization, we model it with symbolic executable verifiers that filter invalid actions before execution.

Based on the trajectory dataset $\mathcal{D}$, we extract all pre-action observations, reconstructed scene graphs, actions, and next observations to construct a global transition pool. Concretely, before each action, we build an agent-visible scene graph $g_t=\Gamma(h_t)$ from the interaction history, which is a lightweight environment-specific structured representation that records only the entities, relations, and interface affordances revealed by the trajectory itself (detailed in Appendix~\ref{app:scene_representation}). Here, $\Gamma(\cdot)$ denotes a deterministic reconstruction operator that only uses information available under the POMDP setting:
\begin{equation}
    \mathcal{Z} = \bigcup_{i=1}^{50} \left\{ z_t^{(i)} = \left(o_t^{(i)}, g_t^{(i)}, a_t^{(i)}, o_{t+1}^{(i)}\right) \right\}_{t=0}^{T_i-1}.
\end{equation}

Based on whether the subsequent observation $o_{t+1}$ indicates a successful action execution, we partition the transition pool $\mathcal{Z}$ into a positive set $\mathcal{Z}^{+}$ and a negative set $\mathcal{Z}^{-}$. Formally, we define an indicator function $\text{Valid}(o)$ that returns $1$ if the observation reflects a valid execution, and $0$ if it indicates an execution failure. The positive and negative sets are then defined respectively as:
\begin{equation}
    \mathcal{Z}^{+} = \{ (o, g, a, o') \in \mathcal{Z} \mid \text{Valid}(o') = 1 \}, \quad \mathcal{Z}^{-} = \{ (o, g, a, o') \in \mathcal{Z} \mid \text{Valid}(o') = 0 \}.
\end{equation}

Next, we perform rule induction on the negative set $\mathcal{Z}^{-}$. The Inductor Agent contrasts the contexts of positive and negative examples to summarize the natural language constraints responsible for action failures, which are then compiled into executable Python verification functions $f_r$. Each rule takes the current observation $o_t$, the reconstructed scene graph $g_t$, and a candidate action $a_t$ as input, outputting a legality decision, an error message, and a revision suggestion:
\begin{equation}
    f_r(o_t, g_t, a_t) \rightarrow (c_r, m_r, u_r).
\end{equation}
Here, $o_t$ is the raw pre-action observation, and $g_t=\Gamma(h_t)$ denotes an agent-visible structured scene graph reconstructed from the interaction history. Moreover, $c_r \in \{0,1\}$ indicates whether the rule permits the action, $m_r$ provides specific error feedback, and $u_r$ offers a targeted correction suggestion. This design not only intercepts explicit errors but also provides the LLM with highly interpretable, closed-loop corrective signals.

Because Feasibility Memory serves as a hard symbolic filter rather than a soft preference signal, our verification protocol is deliberately conservative. We therefore conduct automated verification and filtering of the candidate rules across the entire transition pool $\mathcal{Z}$. First, to prevent the false rejection of genuinely feasible interactions, any rule that incorrectly blocks a positive example is strictly discarded. Subsequently, among all zero-false-rejection rules, we apply a greedy selection strategy based on their coverage of the negative set $\mathcal{Z}^{-}$. For any given rule $r$, the subset of negative examples it successfully intercepts is denoted as:
\begin{equation}
    \text{Cover}(r) = \{ (o, g, a, o') \in \mathcal{Z}^- \mid f_r(o, g, a) = (0, m_r, u_r) \}.
\end{equation}
The system iteratively and greedily selects the rule that covers the maximum number of previously uncovered negative examples, until a predefined rule budget or marginal gain threshold is reached. The final retained set of rules constitutes the Feasibility Memory, denoted as $\mathcal{M}_f$. In this way, fragmented failure experiences are transformed into verifiable, interpretable, and directly executable local symbolic constraints. By assigning feasibility alignment to a rule-based mechanism with explicit logical boundaries, the agent can suppress invalid actions and redundant trial-and-error without sacrificing the neural flexibility required for progress reasoning.

\subsection{Progress Alignment via Neural Memory}
\label{progress}

The objective of Progress Memory is to mitigate global Progress Drift by anchoring the agent to the current semantic stage of the task. Since progress alignment relies on fuzzy, context-dependent semantic generalization rather than strict logical verification, we model it with a neural memory distilled from successful trajectories.

Based on the trajectory dataset $\mathcal{D}$, we filter out all trajectories that successfully completed the task to form the positive experience set:
\begin{equation}
    \mathcal{D}^+ = \left\{ \left(x^{(i)}, o_0^{(i)}, a_1^{(i)}, o_1^{(i)}, \dots, a_{T_i}^{(i)}, o_{T_i}^{(i)}, y^{(i)}\right) \in \mathcal{D} \;\middle|\; y^{(i)} = 1 \right\}.
\end{equation}

Successful trajectories are privileged signals for progress alignment because they reveal which high-level semantic stages actually lead to task completion. Given any successful trajectory $\tau^+ \in \mathcal{D}^+$, we introduce a Distiller Agent to decouple it along the temporal dimension into a task-level procedural blueprint. This blueprint consists of a strictly ordered sequence of progress anchors, denoted as $p_1 \rightarrow p_2 \rightarrow \dots \rightarrow p_L$, where each anchor $p_\ell$ corresponds to a key semantic node in the task progression. To further align high-level semantics with low-level execution patterns, the system synchronously extracts a continuous action chunk corresponding to each anchor from the original trajectory, defined as $c_\ell = \{(o_i, a_i)\}_{i=b_\ell}^{e_\ell}$. Through this process, a single successful experience is ultimately structured and represented as:
\begin{equation}
    \mathcal{P} = \left\{ x, \{(p_\ell, c_\ell)\}_{\ell=1}^{L} \right\},
\end{equation}
where $x$ represents the natural language task instruction.

To support semantic retrieval at both the task and stage granularities, we construct a two-level neural indexing architecture over task instructions and progress anchors. Let the task-level embedding and the anchor-level embedding be denoted as $e_x = \phi_x(x)$ and $e_\ell = \phi_p(p_\ell)$ respectively. The update process of Progress Memory can then be formalized as:
\begin{equation}
    \mathcal{M}_p \leftarrow \mathcal{M}_p \cup \left\{ \left( e_x, \{(e_\ell, c_\ell)\}_{\ell=1}^{L} \right) \right\}.
\end{equation}

This design preserves stage-level task structure while allowing semantic transfer across tasks. Instead of using full trajectories as coarse few-shot examples, Progress Memory retrieves anchor-aligned demonstrations matched to the current stage, providing cleaner progress signals and less irrelevant context. As a result, it offers flexible semantic guidance for task advancement and helps prevent Progress Drift.

\subsection{Dual-Alignment Inference via Neuro-Symbolic Memory}
\label{inference}

We combine the two memories in a unified reasoning loop with explicit functional separation. The symbolic pathway is responsible for feasibility alignment by screening out candidate actions that violate environment constraints and returning verifier feedback for refinement. The neural pathway is responsible for progress alignment, including blueprint generation, progress anchoring, and stage transition.

Given a new task $x^\ast$, the system first retrieves the set of semantically most relevant historical blueprints from the progress memory bank $\mathcal{M}_p$, denoted as $\mathcal{R}(x^\ast) = \operatorname{TopK}_{\mathcal{M}_p}(x^\ast)$. The Blueprint Planner Agent takes $x^\ast$ and the retrieved prior blueprints $\mathcal{R}(x^\ast)$ as conditions to generate a structured blueprint for the current task:
\begin{equation}
    \hat{P} = [\hat{p}_1, \hat{p}_2, \dots, \hat{p}_{\hat{L}}],
\end{equation}
where each $\hat{p}_j$ serves as a progress anchor, defining key state nodes for task execution.

During the execution phase, the system maintains a dynamically activated anchor $\hat{p}_j$. At each decision timestep $t$, the system first utilizes $\hat{p}_j$ to extract a reference action chunk $c_j^\star$ matching the current stage from $\mathcal{M}_p$:
\begin{equation}
    c_j^\star = \operatorname{Retrieve}_{\mathcal{M}_p}(\hat{p}_j).
\end{equation}

The Actor Agent synthesizes the historical observation $h_t$, the original task $x^\ast$, the current anchor $\hat{p}_j$, and the reference action $c_j^\star$ to generate a candidate action $\tilde{a}_t$. Crucially, this neural proposal stage focuses on generating progress-consistent actions rather than hard executability checking. Before execution, the symbolic Feasibility Memory first reconstructs the agent-visible scene graph $g_t=\Gamma(h_t)$ and then acts as an interception module that conducts feasibility verification and iterative refinement on the candidate action:
\begin{equation}
    \tilde{a}_t = \pi_\theta(h_t, \hat{p}_j, c_j^\star), \qquad a_t = \operatorname{Refine}(\pi_\theta, h_t, g_t, \hat{p}_j, c_j^\star, \mathcal{M}_f)
\end{equation}
where $\operatorname{Refine}(\cdot)$ denotes the iterative generation process under symbolic rules and feasibility constraints. The system repeats this process until it either generates an action $a_t$ free of physical and logical violations, or reaches a predefined iteration limit. This ensures the action is strictly grounded in the local environment.

After executing action $a_t$ and obtaining the new observation $o_{t+1}$, the Progress Monitor Agent $\psi_\phi$ evaluates the completeness of the current stage and drives the evolution of the anchor state:
\begin{equation}
    u_t = \psi_\phi(\hat{p}_j, a_t, o_{t+1}, h_t), \qquad j_{t+1} = j_t + u_t
\end{equation}
Here, $u_t \in \{0, 1\}$ is a binary switching signal. When $u_t=1$, the system determines the current anchor task is completed and automatically steps forward to the next progress anchor $\hat{p}_{j+1}$; otherwise, it maintains the current anchor. By routing global progress reasoning to neural memory and local executability checking to symbolic memory, the agent avoids the two characteristic failure modes of single-paradigm systems: semantic drift from under-structured progress modeling and invalid trial-and-error from under-grounded action generation. The result is a dual-alignment loop with decoupled global progress guidance and local feasibility control, jointly preserving forward momentum and local validity.

\vspace{-2mm}
\section{Experiments}
\label{sec:experiments}

\subsection{Experimental Setup}

We evaluate our method on three representative long-horizon agent benchmarks spanning embodied interaction, web-based decision making, and compositional synthesis. (1) \textbf{ALFWorld} \citep{alfworld} is a text-based embodied household environment aligned with ALFRED, where the agent must complete high-level goals through navigation, search, pick-and-place operations, and state-changing actions such as cleaning, heating, and cooling. We follow the standard unseen split and report results on 134 test tasks covering six task types. (2) \textbf{WebShop} \citep{webshop} simulates an e-commerce website, where the agent must navigate webpages, filter product attributes, and make purchase decisions based on natural language shopping intents. We evaluate on 100 tasks and report both success rate and score. (3) \textbf{TextCraft} \citep{adapt} is a Minecraft-style text-based crafting environment in which tasks typically require the agent to recursively construct intermediate materials before producing the target item. It therefore provides a systematic testbed for compositional reasoning and long-chain dependency handling. We evaluate on 100 tasks and report task success rate.

\vspace{-3mm}
\paragraph{Competing Baselines.}
We compare against several representative long-horizon agent methods. (1) \textbf{ReAct} \citep{react} interleaves step-by-step reasoning with action execution. (2) \textbf{Reflexion} \citep{reflexion}improves subsequent attempts through linguistic reflection over failed trajectories. (3) \textbf{ADaPT} \citep{adapt} performs hierarchical task decomposition on demand when direct execution fails. (4) \textbf{StateAct} \citep{stateact} enhances a base agent with explicit state tracking and goal-reinforced self-prompting. (5) \textbf{ExpeL} \citep{expel} extracts reusable experience rules and skills from offline trajectories. (6) \textbf{WALL-E 2.0} \citep{walle2} aligns a neuro-symbolic world model with the environment. (7) \textbf{AWM} \citep{awm} abstracts reusable workflow patterns from successful trajectories into text-based memory. All methods use the same backbone model, gpt-4o-2024-11-20, with temperature set to 0. To prevent test leakage and ensure fairness, methods requiring experience distillation collect offline trajectories using gpt-4o-2024-11-20 on 50 disjoint training tasks to build their reflections or memories. Further details are provided in the appendix \ref{app:supp}.

\vspace{-3mm}
\paragraph{Ablation Setting.}
We evaluate the overall contribution of our two memory modules on the full ALFWorld test set. Additionally, for finer-grained design ablations, we utilize a fixed 50-task subset of ALFWorld to control evaluation cost and enable detailed analysis.

\subsection{Main Results}
\begin{table*}[t]
\centering

\renewcommand{\arraystretch}{1.15}
\setlength{\tabcolsep}{6pt}
\begin{tabular}{lcccc}\toprule 
\textbf{Method} & \textbf{ALFWorld} & \multicolumn{2}{c}{\textbf{WebShop}} & \textbf{TextCraft} \\
\cmidrule(lr){2-2} \cmidrule(lr){3-4} \cmidrule(lr){5-5}
 & \makecell{\textbf{Success} \\ \textbf{Rate (\%)}} & \makecell{\textbf{Success} \\ \textbf{Rate (\%)}} & \textbf{Score}  & \makecell{\textbf{Success} \\ \textbf{Rate (\%)}} \\
\midrule
ReAct\citep{react}  & 76.87 & 32 & 0.5010 & 62 \\
Reflexion\citep{reflexion}  & 82.66 & 35 & 0.5204 & 69 \\
ADaPT\citep{adapt}  & 72.39 & 32 & 0.5355 & 77 \\
StateAct\citep{stateact}  & 63.43 & 17 & 0.2973 & 68 \\
ExpeL\citep{expel} &  85.07 & 29 & 0.4582 & 88 \\
WALL-E 2.0\citep{walle2} & 82.84 & 34 & 0.5998 & 66 \\
AWM \citep{awm} & 88.81 & 32 & 0.5160 & 66 \\
\textbf{Ours} & \textbf{94.78} & \textbf{51} & \textbf{0.7132} & \textbf{94} \\
\bottomrule
\end{tabular}
\caption{\textbf{Main Results.} We compare our proposed dual-alignment framework against competitive baselines on ALFWorld, WebShop, and TextCraft. Our method achieves consistent and substantial improvements in both success rates and task scores across all domains.}
\vspace{-3mm}
\label{tab:main_results}
\end{table*}
Table \ref{tab:main_results} shows that our method consistently outperforms all baselines across three long-horizon benchmarks with different sources of difficulty. On ALFWorld, it improves the success rate from 88.81\% under AWM to 94.78\%. On WebShop, it raises the success rate from 35\% under Reflexion to 51\%, while also improving the score from 0.5998 under WALL-E 2.0 to 0.7132. On TextCraft, it increases the success rate from 88\% under ExpeL to 94\%. More importantly, the strongest prior baseline differs by domain, suggesting that existing methods address only part of the long-horizon challenge. In contrast, our gains remain stable across embodied manipulation, web-based decision making, and compositional planning, indicating that jointly modeling progress alignment and feasibility alignment improves both final task completion and overall decision quality.
\vspace{-3mm}

\subsection{Ablation Studies}
\vspace{-2mm}
We conduct an ablation study to systematically investigate three key questions: (1) whether \textit{Progress Memory} and \textit{Feasibility Memory} provide complementary benefits; (2) whether the performance gains of \textit{Progress Memory} arise from successful experiences per se or from their structured, stage-wise organization; and (3) which form of feasibility constraint best balances local error correction with sustained task progress.

We report three evaluation metrics throughout: success rate (SR), invalid action rate (IAR), and average trajectory length (ATL).

\vspace{-3mm}
\paragraph{Do the two memories provide complementary gains?} We first remove the two memory modules from the full test set to examine whether they address distinct failure modes. Table \ref{tab:ablation_results} shows a clear division of roles. Removing Feasibility Memory causes the largest increase in IAR, from 11.81\% to 26.33\%, and reduces SR to 85.82\%, indicating that a substantial portion of failures comes from locally invalid actions that need to be corrected before execution. In contrast, removing Progress Memory keeps IAR relatively controlled at 12.98\%, but increases ATL from 14.60 to 20.30 and still lowers SR by 4.48 points, suggesting that the agent can remain locally valid while losing efficient stage-level advancement. The full model performs best on all metrics, showing that feasibility alignment mainly prevents local breakdowns, whereas progress alignment mainly sustains global task completion; the two memories are therefore complementary rather than redundant.

\begin{table*}[t]
\centering

\renewcommand{\arraystretch}{1.15}
\setlength{\tabcolsep}{6pt}
\begin{tabular}{lccc}\toprule
\textbf{Method} & \textbf{SR $\uparrow$} & \textbf{IAR $\downarrow$} & \textbf{ATL $\downarrow$} \\
\midrule
Ours w/o both memories & 76.87 & 16.73\%  & 22.57 \\
Ours w/o Progress Memory & 90.30 & 12.98\%  & 20.30 \\
Ours w/o Feasibility Memory & 85.82 & 26.33\%  & 20.49 \\
\textbf{Ours} & \textbf{94.78} & \textbf{11.81\%} & \textbf{14.60} \\
\bottomrule
\end{tabular}
\caption{\textbf{Ablation study on the components of our dual-alignment framework.} Evaluated on the full ALFWorld test set. Removing either Progress Memory or Feasibility Memory leads to performance degradation. The full system achieves the highest Success Rate (SR) and lowest Invalid Action Rate (IAR) alongside optimized Average Trajectory Length (ATL), demonstrating the complementary roles of the two memory modules.}
\vspace{-2mm}
\label{tab:ablation_results}
\end{table*}

\vspace{-3mm}
\paragraph{Does progress gain come from retrieval or from structure?} To isolate the impact of Progress Memory's design, we evaluate it on the 50-task ALFWorld subset while keeping the Feasibility Memory fixed as the default executable verifier. Table \ref{tab:progress_setup_results} shows that the main gain comes from structure, while retrieval is useful only when conditioned on that structure. Standard RAG underperforms even the No Memory baseline, suggesting that task-level retrieval of whole successful trajectories can distract the agent when the retrieved context does not match its current execution stage. In contrast, adding a procedural blueprint raises SR to 92 and sharply reduces ATL from 21.64 to 16.42, indicating that explicit stage decomposition is what primarily improves long-horizon control. Replacing task-level retrieval with anchor-level retrieval further improves SR to 94 and ATL to 16.18, showing that retrieval becomes effective only when it is aligned with the current subgoal. Overall, Progress Memory helps not because it stores more past experience, but because it organizes that experience into stage-aware guidance.

\begin{table*}[t]
\centering

\renewcommand{\arraystretch}{1.15}
\setlength{\tabcolsep}{6pt}
\begin{tabular}{lccccc}\toprule
\textbf{Progress Setup} & \textbf{Blueprint} & \textbf{Success Demos} & \textbf{Retrieval} & \textbf{SR $\uparrow$} & \textbf{ATL $\downarrow$} \\
\midrule
No Memory & $\times$ & $\times$ & $\times$ & 90 & 21.22 \\
Standard RAG & $\times$ & $\checkmark$ & Task & 84 & 21.64 \\
Blueprint + Task & $\checkmark$ & $\checkmark$ & Task & 92 & 16.42 \\
Blueprint + Anchor & $\checkmark$ & $\checkmark$ & Anchor & \textbf{94} & \textbf{16.18} \\
\bottomrule
\end{tabular}
\caption{\textbf{Performance impact of different Progress Memory configurations.} The results demonstrate that standard task-level retrieval is insufficient; combining structured procedural blueprints with anchor-level retrieval yields the highest success rate and best efficiency.}
\vspace{-4mm}
\label{tab:progress_setup_results}
\end{table*}
\vspace{-4mm}


\begin{wraptable}{r}{0.48\textwidth} 
\centering
\footnotesize 

\renewcommand{\arraystretch}{1} 
\setlength{\tabcolsep}{4pt} 

\begin{tabular}{lccc}\toprule
\textbf{Feasibility Setup} & \textbf{SR $\uparrow$} & \textbf{IAR $\downarrow$}  & \textbf{ATL $\downarrow$} \\
\midrule
No Rules & 88 & 21.63\%  & 20.90 \\
Prompt Rules & 84 & 12.11\% & 18.16 \\
Verifier Rules & \textbf{94} & 11.00\%  & \textbf{16.18} \\
Prompt + Verifier & 92 & \textbf{4.66\%} & 16.32 \\
\bottomrule
\end{tabular}
\caption{\textbf{Comparison of different Feasibility Memory constraints.} Compared to purely prompt-based constraints which can lead to overly conservative behavior, executable Verifier Rules achieve the best balance between reducing local invalid actions (IAR) and maintaining continuous global progress (SR and ATL).}
\vspace{-2mm}
\label{tab:feasibility_setup_results}
\end{wraptable}
\vspace{-1mm}
\paragraph{Which feasibility mechanism gives the best trade-off?} We evaluate various constraints on the 50-task ALFWorld subset, fixing Progress Memory to the default blueprint and anchor-level retrieval. Table \ref{tab:feasibility_setup_results} shows that the best feasibility mechanism is not the one that minimizes IAR, but the one that preserves progress. Prompt Rules sharply reduce IAR yet yield the lowest SR, suggesting that a single language prompt is not well suited to handle both semantic task advancement and strict feasibility enforcement. Adding Prompt Rules on top of the Verifier lowers IAR further to 4.66\%, but still underperforms Verifier Rules alone in SR and ATL, indicating that stricter prompt-level filtering does not improve long-horizon control. In contrast, executable Verifier Rules achieve the best SR and ATL with a low IAR, supporting our claim that feasibility alignment should be decoupled from language-based progress guidance and handled by a separate mechanism.

\vspace{-4mm}
\section{Conclusion}
\label{sec:conclusion}
\vspace{-3mm}
This paper addresses the dual-alignment challenge in long-horizon agents by showing that progress alignment and feasibility alignment are best handled by separate mechanisms. We instantiate this insight in a Neuro-Symbolic Dual Memory Framework, where a neural Progress Memory provides stage-aware guidance and a symbolic Feasibility Memory performs executable action verification. Experiments on ALFWorld, WebShop, and TextCraft show that this decoupled design consistently improves task success while reducing invalid actions and redundant interaction, suggesting that stable long-horizon control benefits from matching distinct failure modes with distinct mechanisms. Like current neuro-symbolic agents, the framework requires a certain amount of offline trajectory data for memory construction. In environments with extremely sparse rewards or hard-to-interpret failure signals, this requirement can be difficult to satisfy, making such scenarios an important direction for future work.

\bibliography{colm2026_conference}
\bibliographystyle{colm2026_conference}

\clearpage
\setcounter{page}{1}
\appendix

\section{LLM Usage}
We utilized Large Language Models (LLMs) to assist with the drafting and linguistic refinement of this manuscript. The LLM was employed to optimize language expression, enhance readability, and ensure clarity across key sections, supporting tasks such as sentence restructuring, grammar checking, and improving textual flow.

Notably, the LLM was not involved in the conceptualization, methodology design, or experimental planning of this study. All research concepts, analyses, and conclusions were independently developed by the authors, with the LLM’s contributions strictly limited to linguistic improvement.

The authors take full responsibility for all content, including LLM-assisted text. We confirm that such content complies with ethical guidelines and avoids plagiarism or academic misconduct.

\section{Implementation Details}
\label{app:supp}

\subsection{Training and Evaluation Task Sets}

For all methods that require offline experience collection or distillation, we construct a small training pool that is strictly disjoint from the evaluation tasks. In ALFWorld, we use 50 tasks from the official training split for experience collection. In WebShop and TextCraft, we randomly sample 50 tasks for each environment and ensure that these sampled tasks are fully disjoint from the corresponding evaluation set.

For evaluation, we follow the standard unseen split in ALFWorld and report results on all 134 unseen tasks. For WebShop and TextCraft, following prior works \citep{adapt,stateact}, we evaluate on 100 tasks for each environment, and all evaluation tasks are fully disjoint from the training tasks used for experience collection.

\subsection{Model Configuration}

Unless otherwise specified, all agent roles used in both the training and evaluation stages employ the same backbone model, \textbf{gpt-4o-2024-11-20}. This includes the Inductor Agent, Distiller Agent, Blueprint Planner Agent, Progress Monitor Agent, and Actor Agent. Consistent with the main experiments, all LLM calls use temperature 0.

\subsection{Inference Step Budgets}

We use fixed interaction budgets for each environment. For all methods except ADaPT, the interaction horizon is capped at 50 steps in ALFWorld, 15 steps in WebShop, and 40 steps in TextCraft. ADaPT is the only exception because it performs recursive subtask decomposition: instead of imposing a global trajectory budget, we cap each decomposed subtask to at most 50 steps in ALFWorld, 15 steps in WebShop, and 40 steps in TextCraft, without introducing an additional cap on the total number of steps across the full decomposed trajectory.

\subsection{Retrieval Configuration}

For dense retrieval in Progress Memory, we use cosine similarity as the retrieval metric and \texttt{all-mpnet-base-v2} as the text embedder. For task-level retrieval used to construct the current blueprint, we retrieve the top 3 similar tasks by default. For anchor-level retrieval, we retrieve the top 3 similar anchors in both ALFWorld and WebShop. In TextCraft, however, we do not use anchor-level retrieval (i.e., top 0 anchors), because the crafting dependencies and synthesis rules vary substantially across tasks, making anchor-level transfer much less reliable.

\subsection{Environment-Specific Definition of Valid Signals}

The validity indicator $\text{Valid}(o)$ introduced in Section~\ref{feasibility} strictly captures \emph{explicit environmental rejection} rather than a general lack of task progress. An action is deemed invalid if and only if the environment refuses to execute it or fails to recognize it in the current state. Consequently, actions that are executable but sub-optimal for the global goal are not penalized as "invalid." This decoupling is crucial: global inefficiencies are captured by downstream metrics like Success Rate (SR) and Average Trajectory Length (ATL), whereas the Invalid Action Rate (IAR) is specifically designed to isolate local executability failures. To ensure a strictly fair comparison, this exact environment-specific predicate is applied uniformly across both the rule induction phase (constructing $\mathcal{Z}^{+}$ and $\mathcal{Z}^{-}$) and the evaluation phase for all baseline agents.\begin{itemize}\item \textbf{ALFWorld.} An action is recorded as invalid if the observation is exactly \texttt{"Nothing happens."}, serving as our canonical rejection signal during transition extraction.\item \textbf{WebShop.} Invalid transitions are triggered by the exact observation \texttt{"Invalid action!"}. To minimize sensitivity to superficial wording variations, our WebShop wrapper intentionally normalizes all invalid UI operations to this canonical string, while preserving fine-grained failure reasons in structured metadata.\item \textbf{TextCraft.} Rejections are identified when the observation begins with \texttt{"Could not"}. Additionally, our parser explicitly aligns successful executions with canonical patterns like \texttt{"Got ..."} and \texttt{"Crafted ..."}, ensuring that malformed commands are also strictly categorized as invalid before entering the negative experience pool.\end{itemize}By coupling the validity proxy directly to the environment's native rejection semantics, our framework remains agnostic to arbitrary task-level failures. This design ensures high modularity: if an environment updates the surface form of its error messages, only this thin adapter requires modification, leaving the rule induction objective and evaluation protocol completely intact.

\subsection{Environment-Specific Scene Representation Construction}
\label{app:scene_representation}

The ``scene graph'' in Section~\ref{feasibility} should be understood as an agent-visible structured scene representation. Its concrete form is environment-specific, but in all cases it is deterministically reconstructed from the interaction trajectory available under the POMDP setting, rather than from any latent simulator state or privileged oracle signal.

\paragraph{ALFWorld.}
We maintain an explicit relational scene graph over discovered locations, objects, and location-connectivity edges. The graph is initialized from the set of reachable locations mentioned in the first observation, marks unexplored containers or receptacles, updates the current location and bidirectional edges after successful \texttt{go to} actions, adds newly observed objects when the environment reveals ``you see'' descriptions, and updates object positions after successful \texttt{take}/\texttt{put} interactions. In parallel, we preserve the first observed placement of each object as an \texttt{initial\_state} snapshot and derive a compact symbolic state containing the target item, reachable locations, item placements in currently observed locations, the item in hand, and the agent's current position. This representation supports the ALFWorld verifier rules that check hand occupancy, reachability, and receptacle-usage constraints.

\paragraph{WebShop.}
WebShop uses a lightweight UI-oriented scene graph rather than a full DOM graph. Each snapshot stores a \texttt{page} block with the current page type, query string, page number, product ASIN, subpage, and selected options; a \texttt{ui} block with currently visible clickable targets, result ASINs, and option types; and a \texttt{history} block with visited products, recently clicked targets, and invalid actions. For transition indexing, we additionally derive a minimal pre-action structured state containing only the current page type and visible clickables. This is sufficient for the retained verifier rules, which mainly determine whether \texttt{search} is legal on the current page and whether a requested click target is actually visible.

\paragraph{TextCraft.}
TextCraft does not require an explicit topological object graph, so its scene representation is a deterministic symbolic state summarizing the current crafting context. Specifically, each pre-action state contains the parsed goal, the full recipe list extracted from the task description, the set of craftable items implied by these recipes, the current inventory summary, and an \texttt{inventory\_known} flag. The inventory is updated only from trajectory-visible evidence, namely explicit \texttt{inventory} observations and successful \texttt{get}/\texttt{craft} actions. This representation allows the verifier to check recipe-output consistency and ingredient-availability constraints before executing a \texttt{get} or \texttt{craft} action.

\section{Rule Induction and Filtering Details}

The rule pipeline follows the same three-stage pattern in all environments. First, we convert buffered trajectories into transition triples containing the pre-action structured state, the candidate action, and the binary execution result. Second, the Inductor Agent converts clustered negative transitions into natural-language failure rules and executable Python verifier functions. Third, every candidate verifier is evaluated against the entire positive and negative transition pool before being admitted to the final memory.

Our verification protocol is deliberately conservative. A candidate rule is immediately discarded if it rejects even one positive transition, since such a false rejection would suppress a genuinely executable action at test time. Only zero-false-rejection rules are allowed to compete for coverage on the negative pool. Among those surviving rules, we then apply a greedy set-cover procedure: at each round, we select the rule that blocks the largest number of still-uncovered negative transitions, remove those covered negatives from the uncovered set, and continue until no remaining rule adds new coverage. The selected rules are saved in \texttt{pruned\_rules\_code.json} and loaded directly by the inference-time controller.

The concrete state interfaces differ slightly by environment, but in all cases they are constructed only from the interaction trajectory available to the agent under the POMDP setting, rather than from any latent simulator state or privileged oracle signal. In ALFWorld, each transition additionally includes a scene-graph snapshot reconstructed from the accumulated interaction trace, and the verifier mainly targets object-holding, reachability, and receptacle-usage constraints. In WebShop, the structured state stores the current page type and the set of clickable UI targets observed on the current page, which is sufficient for most invalid search and click actions. In TextCraft, the verifier state contains the parsed goal, recipe list, craftable items, and inventory summary extracted from the task description and interaction feedback, which allows the rules to check recipe-output consistency and input-availability constraints before a \texttt{get} or \texttt{craft} action is executed.

Table \ref{tab:rule_filtering_stats} summarizes the environment-wise statistics of rule verification and pruning.

\begin{table*}[t]
\centering
\small
\renewcommand{\arraystretch}{1.1}
\setlength{\tabcolsep}{6pt}
\resizebox{\textwidth}{!}{%
\begin{tabular}{lccccc}
\toprule
\textbf{Environment} & \textbf{Positive Transitions} & \textbf{Negative Transitions} & \textbf{Candidate Rules} & \textbf{Zero-FP Rules} & \textbf{Selected Rules} \\
\midrule
ALFWorld & 644 & 159 & 17 & 15 & 6 \\
WebShop & 190 & 50 & 8 & 8 & 2 \\
TextCraft & 497 & 219 & 7 & 7 & 3 \\
\bottomrule
\end{tabular}
}
\caption{\textbf{Environment-wise statistics of rule verification and pruning.} Positive transitions are used to eliminate false-positive rules, and negative transitions are used for greedy coverage-based selection. The final rule bank used at inference time is intentionally compact.}
\label{tab:rule_filtering_stats}
\end{table*}

The final selected rules are semantically interpretable. In ALFWorld, the retained rules mainly encode hand-occupancy constraints and location-reachability conditions for \texttt{put}, \texttt{open}, and \texttt{take}. In WebShop, the final bank collapses to two high-coverage rules: one checks whether a \texttt{search} action is legal on the current page, and the other verifies that a clicked target is actually present in the current clickable set. In TextCraft, the retained rules check whether the requested crafting count matches a valid recipe and whether the required ingredients are available in sufficient quantity.

At inference time, the feasibility memory acts as an action-interception module. When a candidate action is rejected by the verifier, the agent is allowed to refine and re-sample the action at most 5 times before falling back to the last sampled action.

\section{Algorithm of Dual-Alignment Inference}
\begin{breakablealgorithm}
    \caption{Dual-Alignment Inference with Neuro-Symbolic Memory}
    \label{alg:dual_alignment_inference}
    \begin{algorithmic}
    \STATE \textbf{Initialize:} 
    \STATE Task instruction $x$
    \STATE Progress Memory $\mathcal{M}_p$
    \STATE Feasibility Memory $\mathcal{M}_f$
    \STATE History $h_0 \leftarrow \emptyset$
    \STATE Current anchor index $j_0 \leftarrow 1$
    \STATE Current timestep $t \leftarrow 0$
    \STATE Maximum refinement iterations $K \leftarrow 5$
    
    \STATE \(\mathcal{R}(x) \leftarrow \text{TopK}_{\mathcal{M}_p}(x)\)
    \STATE Blueprint \(\hat{P} \leftarrow \text{BlueprintPlanner}(x, \mathcal{R}(x))\)
    \STATE Parse \(\hat{P}\) into progress anchors \([\hat{p}_1, \hat{p}_2, \dots, \hat{p}_L]\)
    
    \WHILE{task is not completed \textbf{and} \( j_t \leq L \)}
        \STATE \( c_{j_t}^* \leftarrow \text{Retrieve}_{\mathcal{M}_p}(\hat{p}_{j_t}) \)
        \STATE Candidate action \( \tilde{a}_t \leftarrow \pi_\theta(h_t, \hat{p}_{j_t}, c_{j_t}^*) \)
        \STATE \( a_t \leftarrow \tilde{a}_t \)
        
        \FOR{iteration $k = 0$ to $K$}
            \IF{\( \text{Verify}(a_t, \mathcal{M}_f) \) is True}
                \STATE break
            \ELSE
                \STATE \( a_t \leftarrow \text{Refine}(\pi_\theta, h_t, \hat{p}_{j_t}, c_{j_t}^*, \mathcal{M}_f) \)
            \ENDIF
        \ENDFOR
        
        \STATE \( o_{t+1}, r_{t+1}, \texttt{done} \leftarrow \text{env.step}(a_t) \)
        \STATE \( h_{t+1} \leftarrow h_t \cup \{(o_t, a_t, o_{t+1}, r_{t+1})\} \)
        \STATE Switching signal \( u_t \leftarrow \psi_\phi(\hat{p}_{j_t}, a_t, o_{t+1}, h_t) \)
        \STATE \( j_{t+1} \leftarrow j_t + u_t \)
        
        \IF{\( \texttt{done} \)}
            \STATE break
        \ENDIF
        
        \STATE \( t \leftarrow t + 1 \)
    \ENDWHILE
    \RETURN \( h_t \)
    \end{algorithmic}
\end{breakablealgorithm}

\section{Prompt Templates}

This appendix inlines the core prompt templates used by the six main components of our framework: Distiller, Blueprint Planner, Progress Monitor, Actor, Inductor, and Verifier. We intentionally group prompts by component rather than by directory so that each module can be read independently. We omit large baseline prompt libraries and auxiliary prompt banks that are not direct prompts of these six components, because including them verbatim in a \texttt{listings}-based box can trigger LaTeX dimension-overflow errors.

\newtcblisting{promptbox}[1]{
    enhanced,
    breakable,
    colback=gray!5,
    colframe=black!70,
    title={#1},
    fonttitle=\bfseries,
    listing only,
    boxrule=0.6pt,
    arc=1mm,
    left=1mm,
    right=1mm,
    top=1mm,
    bottom=1mm,
    before skip=8pt,
    after skip=8pt,
    listing options={
        basicstyle=\ttfamily\small,
        breaklines=true,
        breakatwhitespace=false,
        columns=fullflexible,
        keepspaces=true,
        showstringspaces=false,
        upquote=true
    }
}
\subsection{Distiller}

\begin{promptbox}{\textbf{Distiller / ALFWorld / blueprint Extraction Prompt}}
You are given a household manipulation task and its ideal action trajectory. Your job is to:
1 Identify the key blueprints (subgoals or logical steps) necessary to complete the task.
2 Divide the action trajectory into segments, where each segment corresponds to a blueprint.
3 For each blueprint, list the indices of the actions (from the trajectory) that belong to that blueprint.
Instructions:
- Only consider the actions (ignore the observations) when mapping actions to blueprints.
- Each blueprint should represent a clear, meaningful subgoal within the overall task.
- The output should be a JSON array, where each object contains:
- "blueprint": a concise description of the subgoal.
- "actions": a list of action indices.
- - CRITICAL: Do NOT introduce any objects, appliances, or locations in the blueprints that do not explicitly appear in the TASK. For example, if the TASK is "heat some mug and put it in coffeemachine", do NOT generate a blueprint like "Retrieve the mug from the fridge" because "fridge" is not mentioned in the task.

Example 1:
Input:
- Task: put a egg in microwave.
- Trajectory:
1. go to fridge 1
2. open fridge 1
3. take egg 1 from fridge 1
4. go to microwave 1
5. open microwave 1
6. put egg 1 in/on microwave 1
Output:
[
  {{"blueprint": "Find the egg", "actions": [1, 2, 3]}},
  {{"blueprint": "Go to the microwave", "actions": [4, 5]}},
  {{"blueprint": "Put the egg in the microwave", "actions": [6]}}
]
Example 2:
Input:
- Task: put a clean soapbar in countertop.
- Trajectory:
1. go to sinkbasin 1
2. take soapbar 1 from sinkbasin 1
3. clean soapbar 1 with sinkbasin 1
4. go to countertop 1
5. put soapbar 1 in/on countertop 1
Output:
[
  {{"blueprint": "Pick up the soapbar", "actions": [1, 2]}},
  {{"blueprint": "Clean the soapbar", "actions": [3]}},
  {{"blueprint": "Place the soapbar on the countertop", "actions": [4, 5]}}
]
Input:
Task: {TASK}
Trajectory:
{TRAJECTORY}
Now, please generate the blueprint list and map the actions to each blueprint in the same format:
\end{promptbox}

\begin{promptbox}{\textbf{Distiller / WebShop / blueprint Extraction Prompt}}
You are given a WebShop shopping task and an ideal action trajectory.
Your job:
1) Identify key blueprints (subgoals) to complete the task.
2) Segment the action trajectory into blueprint-aligned groups.
3) For each blueprint, list the indices of actions that belong to that blueprint.

Instructions:
- Only use the ACTIONS (ignore observations).
- blueprints should be concise, high-level, and actionable, but MUST reflect WebShop's navigation patterns.
- Do NOT include blueprint steps that depend on copying specific ASINs or option values; use generic wording.
- WebShop is iterative: a search may not find the right item. The trajectory may repeat cycles like:
  - click[Back to Search] -> search[...]
  - browsing result pages via click[Next >] / click[< Prev]
  - opening one or more candidate product pages and then returning to search to try again
  Write blueprints in a LOOP/STOP form using "until ..." (e.g., "Iteratively search and refine the query until results contain a promising candidate under the price limit").
- Do NOT over-segment repeated iterations. If the trajectory repeats a pattern (search/paginate/open candidate/back/re-search), group the whole loop into 1 blueprint with one "until ..." stopping condition, and assign ALL involved action indices to that blueprint.
- Prefer 2-5 blueprints total. Split only on real phase changes, such as:
  - switching from exploring (search/pagination/open-and-reject) to committing to a final candidate product page
  - switching from verifying details (Description/Features/Reviews/Attributes + < Prev) to selecting options
  - switching from selecting options to purchasing
- Recommended blueprint templates (choose what fits the trajectory):
  1) "Iteratively search and browse results (Back to Search -> search, Next/< Prev) until you reach a promising candidate product page that likely matches constraints"
  2) "Inspect candidate product(s) (open product pages; check Description/Features/Reviews/Attributes; use < Prev to return) until you confirm one product satisfies all constraints (type/attributes/price)"
  3) "Select required options (e.g., color/size/pack/material/length) until the chosen configuration matches constraints and Buy Now is available"
  4) "Purchase the product (Buy Now)"
- Do NOT create a blueprint that is only a think[...] action. If think[...] appears, attach it to the most relevant surrounding blueprint.
- If the trajectory contains any line that is not a valid action (not matching the Action API), ignore it.
- Output MUST be valid JSON.

Action API:
- search[query]
- click[target]
- think[text]

Example 1:
Task: i would like a 3 ounce bottle of bright citrus deodorant for sensitive skin, and price lower than 50.00 dollars
Trajectory:
1. search[3 ounce bright citrus deodorant sensitive skin]
2. click[B078GWRC1J]
3. click[bright citrus]
4. click[3 ounce (pack of 1)]
5. click[Buy Now]
Output:
[
  {"blueprint": "Iteratively search and open a promising product page until you reach a candidate that likely matches constraints (3 ounce, bright citrus, sensitive skin, price < $50)", "actions": [1, 2]},
  {"blueprint": "Select required options until the chosen configuration matches constraints", "actions": [3, 4]},
  {"blueprint": "Purchase the product", "actions": [5]}
]

Example 2:
Task: i need a blue wireless bluetooth headphones, and price lower than 60.00 dollars
Trajectory:
1. search[blue wireless bluetooth headphones]
2. click[B09QKP7XQL]
3. click[blue]
4. click[Buy Now]
Output:
[
  {"blueprint": "Iteratively search and open a promising product page until you reach a candidate that likely matches constraints (blue, wireless, bluetooth, price < $60)", "actions": [1, 2]},
  {"blueprint": "Select required options until the chosen configuration matches constraints", "actions": [3]},
  {"blueprint": "Purchase the product", "actions": [4]}
]

Example 3 (iterative re-search):
Task: buy a 20ft video cable that has aluminum alloy, and price lower than 60.00 dollars
Trajectory:
1. search[20ft video cable aluminum alloy]
2. click[Next >]
3. click[B0XXXXXXX]
4. click[Back to Search]
5. search[20ft HDMI cable aluminum alloy price under 60]
6. click[B0YYYYYYY]
7. click[Buy Now]
Output:
[
  {"blueprint": "Iteratively search and browse results (Back to Search -> search, Next/< Prev, open candidates) until you reach a promising product page that likely matches constraints (20ft, aluminum alloy, price < $60)", "actions": [1, 2, 3, 4, 5, 6]},
  {"blueprint": "Purchase the product", "actions": [7]}
]

Example 4 (verify details then choose options):
Task: i want unscented sunscreen lotion for dry skin, and price lower than 40.00 dollars
Trajectory:
1. search[unscented sunscreen lotion dry skin]
2. click[B0AAAAAAA]
3. click[Attributes]
4. click[< Prev]
5. click[Description]
6. click[< Prev]
7. click[Back to Search]
8. search[unscented sunscreen lotion for dry skin under 40]
9. click[B0BBBBBBB]
10. click[unscented]
11. click[Buy Now]
Output:
[
  {"blueprint": "Iteratively search, open candidates, and refine the query until you find a promising product that likely matches constraints (unscented, dry skin, price < $40)", "actions": [1, 2, 3, 4, 5, 6, 7, 8, 9]},
  {"blueprint": "Select required options until the chosen configuration matches constraints", "actions": [10]},
  {"blueprint": "Purchase the product", "actions": [11]}
]

Input:
Task: {TASK}
Trajectory:
{TRAJECTORY}
Now output the blueprint list in the same JSON format:
\end{promptbox}

\begin{promptbox}{\textbf{Distiller / TextCraft / blueprint Extraction Prompt}}
You are given a TextCraft crafting problem and an ideal action trajectory.
Your job:
1) Identify key blueprints (subgoals) to complete the task.
2) Segment the action trajectory into blueprint-aligned groups.
3) For each blueprint, list the indices of actions that belong to that blueprint.

Instructions:
- Only use the ACTIONS (ignore observations).
- The task text includes "Crafting commands:" (recipes) and "Goal:". Use the crafting commands to understand prerequisites and avoid impossible blueprints.
- blueprints should be concise, high-level, and actionable.
- Prefer 3-8 blueprints total. Split only on real phase changes, e.g.:
  - gathering prerequisites (get ...)
  - crafting intermediate items
  - crafting the final goal item
- Use stable phase-level wording because these blueprint texts will later be used for retrieval:
  - gather-only phase: "Gather materials for <next crafted item/phase> (<item1>, <item2>, ...)"
  - short gather+craft phase: "Gather <primitive item(s)> and craft <intermediate item>"
  - craft phase: "Craft <count> <item> ..." or "Craft <item> using ..."
  - loop phase: "Repeatedly ... until ..."
- Do NOT emit one blueprint per simple ingredient get when those gets belong to the same upcoming craft phase.
- Avoid atomic blueprints like "Get 1 stick", "Get 1 flint", "Get 1 feather" if they should be grouped as one gather phase for the next craft.
- Avoid ad-hoc phrasing like "Get 1 poppy and craft 1 red dye". Prefer the canonical styles above so similar tasks use similar blueprint wording.
- If the trajectory repeats a pattern (e.g., repeated dye->wool crafting), do NOT over-segment; group the whole loop into 1 blueprint and write it in a LOOP/STOP form using "until ...", and assign ALL involved action indices to that blueprint.
- Do NOT create a blueprint that is only an inventory / think step. If inventory/think appears, attach it to the most relevant surrounding blueprint.
- Action indices MUST be 1-based (match the numbered trajectory).
- If the trajectory contains any line that is not a valid action, ignore it.
- Output MUST be valid JSON: output ONLY a JSON array (no extra text).

Action API:
- get <count> <item>
- craft <count> <item> using <count> <item>, <count> <item>, ...
- inventory
- think: <text>

Example (real trajectory):
Task:
Crafting commands:
craft 4 oak planks using 1 oak logs
craft 4 stick using 2 planks
craft 1 white wool using 4 string
craft 1 red dye using 1 poppy
craft 1 red wool using 1 red dye, 1 white wool
craft 1 red banner using 6 red wool, 1 stick

Goal: craft red banner.
Trajectory:
1. get 1 oak logs
2. craft 4 oak planks using 1 oak logs
3. craft 4 stick using 2 oak planks
4. get 24 string
5. craft 1 white wool using 4 string
6. craft 1 white wool using 4 string
7. craft 1 white wool using 4 string
8. craft 1 white wool using 4 string
9. craft 1 white wool using 4 string
10. craft 1 white wool using 4 string
11. get 1 poppy
12. craft 1 red dye using 1 poppy
13. craft 1 red wool using 1 red dye, 1 white wool
14. get 1 poppy
15. craft 1 red dye using 1 poppy
16. craft 1 red wool using 1 red dye, 1 white wool
17. get 1 poppy
18. craft 1 red dye using 1 poppy
19. craft 1 red wool using 1 red dye, 1 white wool
20. get 1 poppy
21. craft 1 red dye using 1 poppy
22. craft 1 red wool using 1 red dye, 1 white wool
23. get 1 poppy
24. craft 1 red dye using 1 poppy
25. craft 1 red wool using 1 red dye, 1 white wool
26. get 1 poppy
27. craft 1 red dye using 1 poppy
28. craft 1 red wool using 1 red dye, 1 white wool
29. craft 1 red banner using 6 red wool, 1 stick
Output:
[
  {"blueprint": "Craft basic materials needed for the banner (planks, stick)", "actions": [1, 2, 3]},
  {"blueprint": "Craft 6 white wool from string", "actions": [4, 5, 6, 7, 8, 9, 10]},
  {"blueprint": "Repeatedly make red dye from poppy and combine with white wool until you have 6 red wool", "actions": [11, 12, 13, 14, 15, 16, 17, 18, 19, 20, 21, 22, 23, 24, 25, 26, 27, 28]},
  {"blueprint": "Craft the red banner", "actions": [29]}
]

Input:
Task (Crafting commands + Goal):
{TASK}
Trajectory:
{TRAJECTORY}
Now output the blueprint list in the same JSON format:
\end{promptbox}

\subsection{Blueprint Planner}

\begin{promptbox}{\textbf{Blueprint Planner / ALFWorld / blueprint Guide Prompt}}
You are a professional planner specializing in breaking down complex tasks into clear, blueprint-driven action guides based on expert examples.
Your instructions:
- Carefully study the provided example(s) and reproduce their style exactly in your answer.
- Match the examples in wording, logic, and step order as closely as possible.
- Do not add, remove, or rephrase steps unless strictly necessary to fit the new task.
- Organize your solution as a sequence of major blueprints, each representing a key stage in accomplishing the task, just as in the examples.
- Each blueprint should be concise and actionable, using the same pattern and phrasing style as the examples.
- Do NOT copy demo-specific instance IDs (e.g., "cabinet 5", "drawer 2") unless the new TASK explicitly mentions them. Prefer generic location types (e.g., "cabinet", "drawer").
- Do NOT generate blueprints for "Open" or "Close" actions (e.g., "Open the cabinet", "Close the microwave"). Skip these steps entirely in the guide.
- CRITICAL: Do NOT introduce any objects, appliances, or locations in the blueprints that do not explicitly appear in the TASK. For example, if the TASK is "heat some mug and put it in coffeemachine", do NOT generate a blueprint like "Retrieve the mug from the fridge" because "fridge" is not mentioned in the task.

{EXAMPLES}
Task: {TASK}
Following the provided style and format, outline a blueprint-based action guide for the given task (no unnecessary explanations).
blueprint action guide:
\end{promptbox}

\begin{promptbox}{\textbf{Blueprint Planner / WebShop / blueprint Guide Prompt}}
You are a professional planner for WebShop tasks.
You break a shopping instruction into a short, blueprint-driven action guide.

blueprint style (must follow):
- Write each blueprint in a LOOP/STOP form using "until ...", matching WebShop's iterative navigation.
- Prefer 2-5 blueprints total. Split only on real phase changes.
- Keep blueprints high-level and actionable (avoid low-level UI noise).
- Do NOT include blueprints that require copying specific ASINs or option values; use generic wording.

WebShop navigation patterns to reflect:
- Iterative search is normal: click[Back to Search] -> search[...] may happen multiple times.
- Browsing results: click[Next >] / click[< Prev] across result pages.
- Inspecting candidates: open product pages; check Description/Features/Reviews/Attributes; use < Prev to return.
- Selecting options: pick color/size/pack/material/length until constraints match.
- Buying: click[Buy Now] ends the episode (score shown).

Recommended blueprint templates (use what fits):
1) "Iteratively search and browse results until you reach a promising candidate product page that likely matches constraints"
2) "Inspect candidate product(s) until you confirm one satisfies all constraints (type/attributes/price)"
3) "Select required options until the chosen configuration matches constraints and Buy Now is available"
4) "Purchase the product (Buy Now)"

Output format:
- Output ONLY a JSON array of blueprint strings (no extra text).

Example(s):
{EXAMPLES}

Task: {TASK}
Output (JSON array only):
\end{promptbox}

\begin{promptbox}{\textbf{Blueprint Planner / TextCraft / blueprint Guide Prompt}}
You are a professional planner for TextCraft crafting tasks.
You break a crafting goal into a short, blueprint-driven action guide.

blueprint style (must follow):
- Carefully study the example task + guide pairs and reproduce their phase granularity and wording.
- The blueprint style must strictly match the provided examples because each blueprint will later be used to retrieve matched demonstrations.
- Prefer 3-8 blueprints total. Split only on real phase changes.
- Each blueprint should be concise, high-level, and actionable.
- blueprints MUST be feasible using ONLY:
  - the provided "Crafting commands" (craft ...)
  - the primitive action "get <count> <item>"
- Use these canonical blueprint styles when they fit:
  - gather-only phase: "Gather materials for <next crafted item/phase> (<item1>, <item2>, ...)"
  - short gather+craft phase: "Gather <primitive item(s)> and craft <intermediate item>"
  - craft phase: "Craft <count> <item> ..." or "Craft <item> using ..."
  - loop phase: "Repeatedly ... until ..."
- Do NOT emit one blueprint per simple `get` if those items belong to the same upcoming craft phase.
- Avoid atomic blueprints like "Get 1 stick", "Get 1 flint", "Get 1 feather"; group them into one gather phase for the next craft.
- Avoid ad-hoc wording like "Get 1 poppy and craft 1 red dye". If you combine gathering and crafting, use the example-library style "Gather ... and craft ...".
- If repetition is needed, write it in a LOOP/STOP form using "until ...", e.g.:
  - "Repeatedly get iron nuggets and craft iron ingots until you have at least 10 iron ingots"
- Do NOT include blueprints that are only "inventory" or "think".

Output format:
- Output ONLY a JSON array of blueprint strings (no extra text).

Example(s):
{EXAMPLES}

Task (Crafting commands + Goal):
{TASK}

Output (JSON array only):
\end{promptbox}

\subsection{Progress Monitor}

\begin{promptbox}{\textbf{Progress Monitor / ALFWorld / blueprint Progress Prompt}}
You are a STRICT blueprint verifier for an ALFWorld agent.
Your core objective is to decide whether the agent has successfully completed the CURRENT blueprint based ONLY on explicit OBSERVATIONS in the recent trajectory. 

Be conservative: ONLY advance when justified by OBSERVATIONS (with one explicit quoted snippet).

Inputs:
Task:
{TASK}

blueprint guide (numbered):
{GUIDE}

Current blueprint: {CUR_NUM}/{NUM} (1-based)
Current blueprint text:
{CUR_blueprint}

Recent trajectory (most recent last). Each step has action and observation:
{TRAJECTORY}

=========================================
DECISION PROCEDURE & CHAIN OF THOUGHT (CRITICAL)
=========================================
You MUST evaluate the current trajectory by following these exact steps. You will document your reasoning in the "thought_process" field of your output JSON.

Step 1: Identify blueprint Type & Target. Determine if the blueprint is Pick up, Go to, Open, Clean/Heat/Cool, or Put. Identify the exact object/container. Check for ordinals (e.g., 'first', 'second') or definite articles (e.g., 'the').
Step 2: Bind Variables (if applicable). If the target is "the <container>" without an ID, scan the trajectory backwards to find the exact instance ID the agent most recently arrived at or observed.
Step 3: Resolve Ordinals (if applicable). If looking for a "first/second/third" object, explicitly list all distinct instance IDs of that object base_type picked up so far in your thought process. 
Step 4: Search for Evidence. Scan the trajectory from NEWEST to OLDEST. Look for ONE SINGLE OBSERVATION that proves completion according to the STRICT COMPLETION CRITERIA below.
Step 5: Cascade Check. Verify that you are ONLY attempting to prove the CURRENT blueprint (index: CUR_NUM - 1). You can ONLY advance 0 or 1 blueprint. Never advance multiple.
Step 6: Final Verdict. Determine `next_blueprint_idx` and extract the exact `evidence` substring.

========================
HARD RULES (MUST FOLLOW)
========================
A) Use ONLY OBSERVATIONS as evidence. Ignore ACTION text except to locate the relevant step.
B) You may advance a blueprint ONLY if you can cite evidence from ONE SINGLE OBSERVATION (one step).
   - Evidence MUST be an EXACT verbatim substring from that OBSERVATION.
C) If you cannot find qualifying evidence for the CURRENT blueprint, you must NOT advance.
D) CASCADE CONTROL (CRITICAL): You are NOT allowed to advance more than ONE blueprint per call.
   - You may either: NOT advance (stay on current blueprint), OR advance EXACTLY ONE blueprint.
E) Object matching is STRICT. No synonyms unless an explicit alias list is provided in GUIDE/TASK.
F) Indexing is critical:
   - current_idx = CUR_NUM - 1 (0-based).
   - Output next_blueprint_idx MUST be 0-based.
   - If you do NOT advance, output next_blueprint_idx = current_idx exactly.
   - If you DO advance, output next_blueprint_idx = current_idx + 1 exactly.
   - You may NEVER output next_blueprint_idx > current_idx when evidence_step = 0.
G) Output ONLY valid JSON. No markdown wrappers. No extra text outside the JSON.
H) Avoid standalone YES/NO/TRUE/FALSE anywhere. Use "proven"/"not proven".

=========================================
NORMALIZATION & STRICT MATCH
=========================================
1) Normalize all names for matching: lowercase, remove articles ("a", "an", "the").
2) [CRITICAL] Treat simulator object types as atomic tokens; DO NOT use real-world knowledge (e.g., "handtowel" is NOT "cloth"). Common non-equivalences:
   - cloth != handtowel != towel != dishsponge != papertowelroll
   - towelholder != handtowelholder
   - mug != cup
3) If an object has a trailing integer id (e.g., "keychain 3"), that is its instance id. base_type("keychain 3") = "keychain".
4) Strict match: If a blueprint target includes an instance id, it must match EXACTLY. If omitted, match ONLY by exact base_type.

=========================================
DEFINITE REFERENCE BINDING ("the X")
=========================================
If a blueprint refers to "the <container>" (e.g., "the cabinet"):
- Find the most recent observation that unambiguously indicates the agent is at an instance of that container:
  - "You arrive at <container> <id>."
  - "On the <container> <id>, you see ..."
  - "The <container> <id> is closed/open." 
- Bind "the <container>" to that exact instance (e.g., "cabinet 1"). For "Go to" blueprints, you MAY use the same observation for both binding and evidence.

=========================================
ORDINAL ITEMS (first/second/third...)
=========================================
If the blueprint includes ordinals, reconstruct the state from the PROVIDED TRAJECTORY (OBS-only):
1) Scan from OLDEST to NEWEST step.
2) Extract pickup events of the base_type with an instance id: "You pick up the <obj> <id> ..."
3) Build a list `unique_picked_ids` by first appearance order (dedupe by id). List this in your thought process.
4) Define first_id (index 0), second_id (index 1), etc.
5) "second <obj>" is completed ONLY if second_id exists AND the evidence OBS shows picking up OBJ with id == second_id. Re-picking the first instance NEVER counts as second.
6) "put the first/second <obj> in/on <Y>" MUST involve the corresponding tracked instance id.

=========================================
blueprint COMPLETION CRITERIA (OBSERVATION PATTERNS)
=========================================
1) Find / Pick up X
   Completed ONLY if OBS explicitly confirms picking up: "You pick up the <obj> ..." OR "You pick up <obj> ..." OR inventory line: "You are carrying: ... <obj> ...". Seeing ("you see <obj>") does NOT count.

2) Go to Y
   Completed ONLY if OBS shows the agent is at Y: "You arrive at <Y>." OR "On the <Y>, you see ..." OR "The <Y> is closed/open."

3) Open CONTAINER (cabinet/fridge/microwave/drawer/door/safe/etc.)
   Apply binding first.
   - EXPLICIT CLOSED: If an OBS about that exact container contains "The <container> is closed.", Open is NOT proven.
   - EXPLICIT OPEN: Completed if OBS contains: "You open the <container>." OR "The <container> is open." OR "The <container> is already open."
   - SIMULATOR SKIP-OPEN RULE (CRITICAL): Find the most recent "at-container" arrival OBS ("You arrive at...", "On the <container>, you see..."). If that arrival OBS does NOT contain the substring "The <container> is closed.", then the blueprint IS COMPLETED (auto-skip). Advance next_blueprint_idx to current_idx+1 and quote the arrival OBS as evidence.

4) Clean / Heat / Cool X
   Completed ONLY if OBS explicitly confirms success: "You clean the X" / "You heat the X" / "You cool the X". Merely being at the appliance does NOT count.

5) Put X in/on Y
   Completed ONLY if OBS explicitly confirms: "You move the <X> to the <Y>." OR "You put the <X> in/on the <Y>."
   NOT EVIDENCE: "In it, you see...", "On the <Y>, you see...", being at Y, opening/closing Y, or inventory lines alone.

========================
OUTPUT FORMAT
========================
Output ONLY a valid JSON object. Do NOT wrap it in markdown formatting like ```json. The "thought_process" MUST be the very first key.

{{
  "thought_process": "Step 1: The current blueprint is ... Step 2: Binding ... Step 3: Ordinals ... Step 4: Scanning for evidence ... Step 5: Cascade check ... Step 6: Conclusion.",
  "next_blueprint_idx": <0-based int>,
  "evidence_step": <1-based index into the recent trajectory, or 0 if not proven>,
  "evidence": "<exact quote snippet, or empty string>",
  "reason": "<short; proven/not proven>"
}}
\end{promptbox}

\begin{promptbox}{\textbf{Progress Monitor / WebShop / blueprint Progress Prompt}}
You are a STRICT blueprint progress checker for a WebShop agent.
Decide whether to advance the current blueprint index based ONLY on the OBSERVATIONS in the recent trajectory.

Inputs:
Task:
{TASK}

blueprint guide (numbered):
{GUIDE}

Current blueprint: {CUR_NUM}/{NUM} (1-based)
Current blueprint text:
{CUR_blueprint}

Recent trajectory (most recent last). Each step has action and observation:
{TRAJECTORY}

=========================================
DECISION PROCEDURE (REQUIRED)
=========================================
You MUST follow these steps and write a concise trace in the "thought_progress" field of your output JSON.

Step 1: Restate the CURRENT blueprint's stopping condition (the "until ..." clause) in your own words.
Step 2: Scan observations from NEWEST to OLDEST and look for ONE SINGLE observation that clearly proves the stopping condition is satisfied.
Step 3: Filter out invalid proof: ignore any observation that starts with "[Rule_" or "Invalid action format", or is exactly "OK.".
Step 4: Cascade control: you are proving ONLY the CURRENT blueprint. You may advance by at most ONE.
Step 5: Finalize: set `next_blueprint_idx`, `evidence_step`, and quote the exact evidence substring.

Hard rules:
1) Use ONLY observations as evidence (quote an exact substring).
2) You may either stay on the current blueprint OR advance by exactly ONE blueprint.
3) Evidence MUST come from ONE SINGLE observation step (no combining across steps).
4) Output JSON ONLY. No markdown. No extra text outside the JSON.
5) The "thought_progress" MUST be the very first key in the JSON.

How to interpret blueprints:
- blueprints are written as "... until ...". You may advance ONLY if the stopping condition ("until ...") is clearly satisfied by an observation.
- Be conservative: if the observation does not clearly prove completion, do NOT advance.
- Ignore non-environment feedback as proof (these do NOT complete any blueprint): observations starting with "[Rule_" or "Invalid action format" or exactly "OK.".

Evidence rubric (helpful but not exhaustive):
- Search / browse / refine loop is proven if an observation shows search results (contains "Page" and "Total results") OR shows a product detail page (contains "Price:" and "[Buy Now]") reached from results.
- Inspect / verify candidates is proven if an observation shows product detail content you can cite (e.g., "Price:", "material", "size", or keywords from the Task such as colors/sizes/materials), typically on Description/Features/Reviews/Attributes pages.
- Select options is proven if an observation indicates a selection, e.g. contains "You have clicked" OR shows a selected marker like "[[...]]" OR the options list changes to reflect a chosen variant.
- Purchase is proven only if an observation contains "Your score (min 0.0, max 1.0)".

========================
OUTPUT FORMAT
========================
Output ONLY a valid JSON object. Do NOT wrap it in markdown formatting like ```json.

{
  "thought_progress": "Step 1: ... Step 2: ... Step 3: ... Step 4: ... Step 5: ...",
  "next_blueprint_idx": <0-based int>,
  "evidence_step": <1-based index into the recent trajectory, or 0 if not proven>,
  "evidence": "<exact quote substring from ONE SINGLE observation, or empty string>",
  "reason": "<short; proven/not proven>"
}

Now output the JSON decision:
\end{promptbox}

\begin{promptbox}{\textbf{Progress Monitor / TextCraft / blueprint Progress Prompt}}
You are a STRICT blueprint progress checker for a TextCraft agent.
Decide whether to advance the current blueprint index based on the recent trajectory and the current inventory.

Inputs:
Task:
{TASK}

blueprint guide (numbered):
{GUIDE}

Current blueprint: {CUR_NUM}/{NUM} (1-based)
Current blueprint text:
{CUR_blueprint}

Current inventory (most up-to-date):
{INVENTORY}

Recent trajectory (most recent last). Each step has action and observation:
{TRAJECTORY}

Decision procedure:
1) Identify the current blueprint's required target and count.
2) Check the inventory line and the recent trajectory from newest to oldest.
3) Find ONE exact quoted snippet that proves the blueprint is completed, or conclude that it is not yet proven.
4) Be conservative: if the evidence is incomplete or ambiguous, do NOT advance.
5) You may either stay on the current blueprint OR advance by exactly ONE blueprint.

Hard rules:
1) Use ONLY environment observations and the inventory line above as evidence.
2) Quote an exact substring for the evidence field. If not proven, evidence must be an empty string.
3) Ignore these as proof of completion: observations exactly "OK.", or observations starting with "Invalid action:" or "Could not".
4) Output JSON ONLY. No markdown. No extra text outside the JSON.
5) The JSON MUST include a short "thought_process" field that summarizes your reasoning steps.

How to interpret blueprints:
- If a blueprint says "... until ...", you may advance ONLY if the stopping condition ("until ...") is clearly satisfied.
- Otherwise, treat a blueprint as complete when its key requirement is clearly satisfied, e.g.:
  - a successful "Got ..." observation for a required item
  - a successful "Crafted ..." observation for a required crafted item
  - the inventory line shows the required item count is reached

Output format:
{
  "thought_process": "Brief step-by-step reasoning. Mention target, checked evidence, and final conclusion.",
  "next_blueprint_idx": <0-based int>,
  "evidence_step": <1-based index in the recent trajectory or 0>,
  "evidence": "<exact quote or empty>",
  "reason": "<proven/not proven>"
}

Now output the JSON decision:
\end{promptbox}

\subsection{Actor}

\begin{promptbox}{\textbf{Actor / ALFWorld / Stage-Conditioned Action Prompt}}
You are controlling ALFWorld. Use the task interaction history exactly as in a ReAct-style trace.

blueprint action guide (high-level plan; do NOT copy object IDs/targets):
{blueprint_ACTION_GUIDE}

Subgoal (current blueprint; treat as a multi-step subtask):
{CURRENT_blueprint}

Retrieved blueprint-level demonstrations (patterns only; do NOT copy IDs/targets):
{blueprint_LEVEL_DEMONSTRATIONS}

Task interaction history (most recent last):
{HISTORY}

Now output the next action (one line only):
>
\end{promptbox}

\begin{promptbox}{\textbf{Actor / WebShop / Stage-Conditioned Action Prompt}}
You are controlling WebShop. Use the task interaction history exactly as in a ReAct-style trace.

blueprint action guide (high-level plan; do NOT copy product IDs/options):
{blueprint_ACTION_GUIDE}

Subgoal (current blueprint; treat as a multi-step subtask):
{CURRENT_blueprint}

Retrieved blueprint-level demonstrations (patterns only; do NOT copy IDs/ASINs/options):
{blueprint_LEVEL_DEMONSTRATIONS}

Task interaction history (most recent last):
{HISTORY}

Now output the next action (one line only):
>
\end{promptbox}

\begin{promptbox}{\textbf{Actor / TextCraft / Programmatically Assembled Action Prompt Template}}
Task (Crafting commands + Goal):
{TASK}

[blueprint Guide]
{blueprint_ACTION_GUIDE}

[Current blueprint]
{CURRENT_blueprint}

[Few-shot Example]
{blueprint_LEVEL_DEMONSTRATIONS}

[Current Trajectory]
{HISTORY}

> 
\end{promptbox}

\subsection{Inductor}

\begin{promptbox}{\textbf{Inductor / ALFWorld / System Prompt}}
You are a rule miner tasked with extracting meaningful rules, improve rules with a collection of transitions in Alfworld. Each transition consists of an initial state, an action, and the action result. 
The rules are for mapping the inputs, 'initial state' and 'action' to their corresponding 'action_result'. 
You need to verify that the given rules satisfy all the transitions, find any conflicting rules and modify them. 
Additionally, try to mine new additional rules from these transitions, new rules must be different from the given rules, and only rules for under what conditions an action will fail need to be generated.

The actions in the transitions are introduced as follows:
go to [location/object]: Move to a specified location or object. 
open [object]: Open a specified object like a cabinet or drawer. 
close [object]: Close an opened object. 
take [object] from [location]: Pick up an item from a specified location. 
put [object] in/on [location]: Place an item in or on a specified location. 
clean [object] with [location/tool]: Clean an object using a specific location or tool, like cleaning lettuce at the sink basin. 
heat [object] with [tool]: Use an appliance, such as a microwave, to heat an item. 
cool [object] with [tool]: Use a cooling tool or appliance, such as a fridge, to cool an item. 
use [tool]: Activate or use a tool, such as a desklamp. 


I will give you an array of transitions:
[
    {
        'initial_state': '...', 
        'action': '...', 
        'action_result': "Whether the action is executed successfully, give 'True' or 'False' only"
    },
    {
        'initial_state': '...', 
        'action': '...', 
        'action_result': "Whether the action is executed successfully, give 'True' or 'False' only"
    },
    ...
]
and an array of rules:
[
    "Rule 1: For action ..., if..., the action will fail; Checking Method: ...",
    "Rule 2: For action ..., if..., the action will fail; Checking Method: ...",
    "Rule 3: For action ..., if..., the action will fail; Checking Method: ...",
    "Rule 4: For action ..., if..., the action will fail; Checking Method: ...",
    ...
]

You should only respond in the format as described below:
RESPONSE FORMAT:
{
    "verified_rules":[
        "Rule ...: ...; Checking Method: ...",
        "Rule ...: ...; Checking Method: ...",
        "Rule ...: ...; Checking Method: ...",
        "Rule ...: ...; Checking Method: ...",
        "Rule ...: ...; Checking Method: ...",
        "Rule ...: ...; Checking Method: ...",
        "Rule ...: ...; Checking Method: ...",
        ...
    ],
    "conflicting_rules":[
        "Rule ...: ...; Checking Method: ...",
        "Rule ...: ...; Checking Method: ...",
        "Rule ...: ...; Checking Method: ...",
        "Rule ...: ...; Checking Method: ...",
        "Rule ...: ...; Checking Method: ...",
        "Rule ...: ...; Checking Method: ...",
        ...
    ],
    "improved_rules":[
        "Rule ...: ...; Checking Method: ...",
        ...
    ],
    "new_rules":[
        "Rule ...: ...; Checking Method: ...",
        ...
    ],
    "final_rules":[
        "Rule ...: ...; Checking Method: ...",
        "Rule ...: ...; Checking Method: ...",
        "Rule ...: ...; Checking Method: ...",
        "Rule ...: ...; Checking Method: ...",
        "Rule ...: ...; Checking Method: ...",
        "Rule ...: ...; Checking Method: ...",
        "Rule ...: ...; Checking Method: ...",
        ...
    ]
}

where
verified_rules: list rules that satisfy all the provided transitions.
conflicting_rules: list rules that contradict any of the transitions. Modify these rules if they can be modified correctly and put them in 'improved_rules'.
improved_rules: show modified 'conflicting_rules'.
new_rules: list new rules discovered. New rules must be different from the rules in 'verified_rules' and 'improved_rules' and satisfy all the transitions. otherwise, simply leave this section blank.
final_rules: combine all the rules from 'verified_rules', 'improved_rules', and 'new_rules'.


Instructions:
- Ensure the response can be parsed by Python `json.loads`, e.g.: no trailing commas, **no single quotes**, etc.
- Please use you knowledge in Alfworld, do inductive reasoning. You need to dig up as many rules as possible that satisfy all transitions.
- Extract and utilize only the features that influence the outcome of the action.
- Please generate general and universal rules; the rules should not reference any specific item or tool! You need to generalize across various items or tools.
- Generate only the rules under what conditions the action will fail.
- While generating a rule, you also need to state how to check if a transition satisfies this rule. Please be specific as to which and how 'state features' in 'initial state' need to be checked
\end{promptbox}

\begin{promptbox}{\textbf{Inductor / ALFWorld / Query Template}}
My information is as follows: 
transitions:
{transitions}
given rules:
{rules}
\end{promptbox}

\begin{promptbox}{\textbf{Inductor / WebShop / System Prompt}}
You are a rule miner tasked with extracting and improving **action feasibility rules** for the WebShop environment.

Each transition is a dictionary with:
- initial_state: a JSON object describing the current page/state (lightweight)
- action: a JSON object describing the proposed action
- action_result: a boolean indicating whether the action executed successfully (True) or was invalid (False)
- sg_info (optional): a scene-graph snapshot capturing UI/page memory (richer than initial_state)

Your job:
1) Verify the given rules against the provided transitions.
2) Fix any conflicting rules (if possible).
3) Mine additional NEW rules.

IMPORTANT:
- Only generate rules for **when an action will fail** (i.e., action_result == False).
- Rules must be **general/universal**. Do NOT reference specific ASINs, product titles, or option values.
- It is allowed to reference fixed UI button names: "Buy Now", "Back to Search", "Next >", "< Prev", "Description", "Features", "Reviews", "Attributes".
- Prefer using the provided state fields to define conditions.

Action API (from transitions):
- search[query]  -> action JSON: {"name":"search","args":{"query":"..."}}
- click[target]  -> action JSON: {"name":"click","args":{"target":"..."}}
- think[text]    -> action JSON: {"name":"think","args":{"text":"..."}}

State format (example):
{
  "page_type": "init|search|item|item_sub|end|unknown",
  "clickables": ["Search", "Back to Search", "B078...", "..."],
  "raw_observation": "..."
}

sg_info format (example; you may use this for more precise rules):
{
  "page": {
    "page_type": "...",
    "query_string": "...",
    "page_num": 1,
    "asin": "B09....",
    "subpage": "Description|Features|Reviews|Attributes|",
    "selected_options": {"color": "blue"}
  },
  "ui": {
    "clickables": [...],
    "asins": [...],
    "option_types": {"blue": "color"}
  },
  "history": {
    "visited_asins": [...],
    "clicked_targets": [...],
    "invalid_actions": [...]
  }
}

Rules should be written in natural language like:
"Rule 1: For action click, if <condition>, the action will fail; Checking Method: <how to check in state/action>"

You MUST respond with a single JSON object (parsable by python json.loads), with keys:
{
  "verified_rules": [...],
  "conflicting_rules": [...],
  "improved_rules": [...],
  "new_rules": [...],
  "final_rules": [...]
}

Guidance:
- Focus on failure causes like: wrong page_type, clicking a target not present in state.clickables, invalid special buttons on wrong pages, empty/invalid search query format, etc.
- Each rule MUST include a concrete Checking Method using state/action fields.
\end{promptbox}

\begin{promptbox}{\textbf{Inductor / WebShop / Query Template}}
My information is as follows:
transitions:
{transitions}
given rules:
{rules}

\end{promptbox}

\begin{promptbox}{\textbf{Inductor / TextCraft / System Prompt}}
You are a rule miner for TextCraft, a simplified Minecraft crafting environment.

You will be given an array of transitions. Each transition has:
- initial_state: JSON object (inventory / allowed recipes / goal)
- action: JSON object (get/craft + arguments)
- action_result: boolean (True = the action executed successfully, False = it failed)

Your task:
1) Verify which given rules are consistent with ALL provided transitions.
2) Fix conflicting rules if possible.
3) Mine NEW additional rules that explain WHEN an action will FAIL.

Important:
- Only generate rules for failure conditions.
- Rules must be general and universal; do NOT reference specific episode seeds.
- Rules must not rely on hidden environment internals.
- The rules should be implementable as Python checks using ONLY `initial_state` and `action`.

Action space:
- inventory
- get <count> <item>
- craft <count> <item> using <count> <item>, <count> <item>, ...

Here is the state/action schema:

initial_state:
{
  "goal": {"item": string, "count": int},
  "recipes": [
    {
      "output": {"item": string, "count": int},
      "inputs": [{"item": string, "count": int}, ...],
      "raw": string
    }
  ],
  "craftable_items": [string, ...],
  "inventory": { "<item>": int, ... },
  "inventory_known": boolean
}

action:
{
  "name": "get" | "craft",
  "args": ...,
  "raw": string
}

get args:
{"count": int, "item": string}

craft args:
{
  "count": int,
  "item": string,
  "inputs": [{"count": int, "item": string}, ...]
}

Rule writing format:
- Each rule MUST be a single string:
  "Rule N: For action <get/craft>, if <condition>, the action will fail; Checking Method: <specific checks on initial_state/action>."

Response format (MUST be valid JSON):
{
  "verified_rules": [...],
  "conflicting_rules": [...],
  "improved_rules": [...],
  "new_rules": [...],
  "final_rules": [...]
}

Guidelines:
- Focus on mechanics: recipe validity, inventory sufficiency, and constraints implied by transitions.
- If `inventory_known` is false, rules should not assume exact counts from inventory.
- Prefer simple and robust conditions that generalize.
\end{promptbox}

\begin{promptbox}{\textbf{Inductor / TextCraft / Query Template}}
My information is as follows:
transitions:
{transitions}
given rules:
{rules}

\end{promptbox}

\end{document}